\documentclass[10pt,twocolumn,a4paper]{ICCAS2024}
 
\usepackage{diagbox}
\usepackage{url}            
\usepackage{booktabs}       
\usepackage{amsfonts}       
\usepackage{nicefrac}       
\usepackage{microtype}      
\usepackage{color}          
\usepackage{graphicx}
\usepackage{booktabs}
\usepackage{caption}
\usepackage{float}
\usepackage{hyperref}
\usepackage[utf8]{inputenc}
\usepackage{algpseudocode}
\usepackage{algorithm}
\usepackage{multirow}

\usepackage{geometry}

\newcommand{\mytoprule}{\specialrule{0.12em}{0em}{0em}}
\newcommand{\mymidrule}{\specialrule{0.05em}{0em}{0em}}
\newcommand{\mybottomrule}{\specialrule{0.12em}{0em}{0em}}

\begin{document}

\title{\textbf{Parallel Distributional Deep Reinforcement Learning\\for Mapless Navigation of Terrestrial Mobile Robots}}

\author{Victor A. Kich${}^{1}$, Alisson H. Kolling${}^{2}$, Junior Costa de Jesus${}^{2}$, Gabriel V. Heisler${}^{3}$, Hiago Jacobs${}^{4}$, Jair A. Bottega${}^{1}$, André L. da S. Kelbouscas${}^{4}$, Akihisa Ohya${}^{1}$, Ricardo B. Grando${}^{2,4}$, Paulo L. J. Drews-Jr${}^{2}$, Daniel F. T. Gamarra${}^{3}$}

\affils{
${}^{1}$ Intelligent Robot Laboratory, University of Tsukuba, (victorkich98@gmail.com) {\small${}^{*}$ Corresponding author}\\
${}^{2}$ Federal University of Rio Grande,\\
${}^{3}$ Federal University of Santa Maria,\\
${}^{4}$ Technological University of Uruguay}


\abstract{
This paper introduces novel deep reinforcement learning (Deep-RL) techniques using parallel distributional actor-critic networks for navigating terrestrial mobile robots. Our approaches use laser range findings, relative distance, and angle to the target to guide the robot. We trained agents in the Gazebo simulator and deployed them in real scenarios. Results show that parallel distributional Deep-RL algorithms enhance decision-making and outperform non-distributional and behavior-based approaches in navigation and spatial generalization.
}


\maketitle


\section{Introduction}
Deep Reinforcement Learning (Deep-RL) has shown significant potential in engineering and robotics for controlling discrete and continuous systems \cite{lillicrap2015continuous}. Initially applied to stable environments \cite{gu2016continuous}, its complexity increases with non-stationary robots like terrestrial mobile robots due to environmental interactions.

To address this, new Deep-RL techniques focusing on action discretization have been developed \cite{zhu2017target}, achieving success in mapless navigation for various mobile robots \cite{de2021soft}, \cite{grando2021deep}, \cite{grando2022double}. Distributed Deep-RL approaches offer promising solutions to the training time issue in limited simulated environments \cite{barth2018distributed}. However, autonomous navigation of terrestrial mobile robots in complex environments remains challenging.

We present two new Deep-RL approaches using parallel distributional techniques: Parallel Distributional Deterministic Reinforcement Learning (PDDRL) and Parallel Distributional Stochastic Reinforcement Learning (PDSRL), incorporating prioritized memory replay for enhanced navigation in complex scenarios. As shown in Fig.~\ref{fig:structure}, our methods use 24-dimensional range findings and the relative distance and angle to the target. Multiple agents' simultaneous learning improves performance in both simulation and real-world scenarios. We used the Turtlebot3 Burger robot for extensive evaluations in four increasingly complex scenarios, including an additional real scenario for spatial generalization testing.

\setcounter{figure}{0}
\begin{figure}[tbp!]
    \centering
    \includegraphics[width=\linewidth]{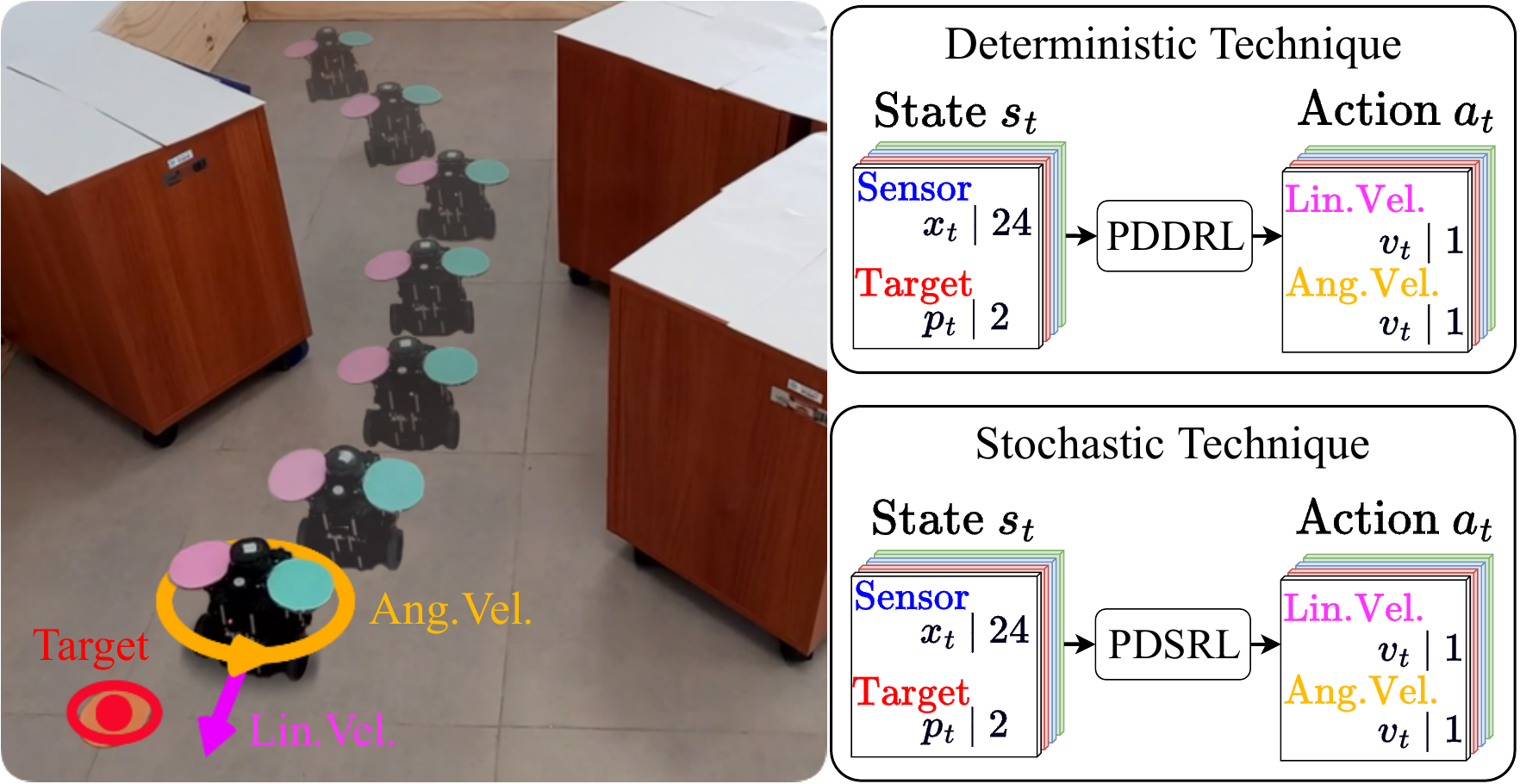}
    \caption{Left: Turtlebot3 Burger navigating a real-world obstacle scenario. Right: Input and output structure of our proposed PDDRL and PDSRL approaches.}
    \label{fig:structure}
    \vspace{-6mm}
\end{figure}

This work's contributions include:
\begin{itemize}
\item Two new distributional Deep-RL approaches for improving goal-oriented mapless navigation using a simple range-based sensing architecture.
\item Demonstration of the feasibility of our approaches through sim-to-real evaluation, addressing challenges such as imprecision and delays.
\item Evidence that the stochastic actor-critic technique with prioritized experience replay outperforms non-distributional techniques and classic algorithms, marking the first extensive sim-to-real evaluation for mapless navigation of terrestrial mobile robots using parallel distributional Deep-RL approaches. \footnote{\scriptsize \url{https://youtu.be/cOVOijEwLUA}}\footnote{\scriptsize \url{https://github.com/victorkich/Parallel-Turtle-DRL}}
\end{itemize}

\section{Related Works}\label{related_works}

Traditional Deep-RL models rewards as a single value, but Bellemare \emph{et al.}~\cite{bellemare2017distributional} proposed modeling it as a probabilistic distribution. This idea was extended to the new implementation of Soft Actor-Critic (SAC) algorithm by Duan \emph{et al.}~\cite{duan2021distributional}, which inspired our approaches.

Distributed Deep-RL, introduced to speed up training, involves distributing computation across multiple processors~\cite{horgan2018distributed}. Mnih \emph{et al.}~\cite{pmlr-v48-mniha16} used asynchronous actor-critic methods, while Horgan \emph{et al.} developed the \textit{Ape-X} method, which decouples the actor from the learner~\cite{horgan2018distributed}. Barth-Maron \emph{et al.}~\cite{barth2018distributed} further improved this with the Deep Deterministic Policy Gradient (DDPG) algorithm. 

Tai \emph{et al.}~\cite{tai2017virtual} demonstrated Deep-RL's application in mobile robotics, inspiring advancements in mapless navigation for terrestrial robots~\cite{zhu2017target}. Jesus \emph{et al.}~\cite{de2021soft} explored deterministic and stochastic approaches in mapless navigation without parallel networks. 

Our work introduces parallel Deep-RL approaches for mapless navigation of terrestrial mobile robots, addressing more complex scenarios with sim-to-real evaluation. We propose PDDRL (deterministic) and PDSRL (stochastic) approaches, compared with the traditional Behavior-based algorithm (BBA) \cite{Lumelsky1987} and parallel versions of DDPG and SAC using both classic and prioritized memory replay.

\section{Methodology}\label{methodology}

In this section, we introduce the central concept of Deep-RL, encompassing the details of both deterministic and stochastic approaches. Additionally, we present our simulated and real scenarios, addressing specific issues such as the rewarding system and the network architecture.

\subsection{Deep Reinforcement Learning}

The objective of standard reinforcement learning is to maximize the expectation of the sum of discounted rewards. The state-action value function is a mathematical model that describes the expected return when taking an action $a$ from a state $s$, and then acting according to the policy $\pi$ \cite{sutton2018reinforcement}. This function, known as
\begin{equation}
    Q_\pi(s,a)=\mathbb{E}\left[\sum_{t=0}^{\infty}\gamma^t r(s_t,a_t)\right],
\end{equation}
is typically utilized to evaluate the quality of a policy.

At each time step, the networks predict the behavior for the present state and generate a signal of temporal difference (TD) error. The Bellman operator 
\begin{equation}\label{basic_bellman}
    (T_\pi Q)(s,a)=r(s,a)+\gamma\mathbb{E}\left[Q(s',\pi(s'))\ |\ s,a\right]
\end{equation}
can minimize this TD error, whose expectation is computed with respect to the next state $s'$.

This work employs two distinct neural networks, namely an actor and a critic, to evaluate the TD error. The TD error is assessed under a separate target policy and value network, where the networks have distinct parameters $(\theta',\omega')$ to stabilize learning. The critic-network generates the Q-value for action, whereas the output of the actor-network is a real value that represents the selected action.

\subsection{Parallel Distributional Deterministic RL}\label{pddrl}

The DDPG architecture, as proposed by \cite{silver2014deterministic} and extended in D4PG \cite{barth2018distributed}, serves as a cornerstone in Deep-RL applications for mobile robots in continuous observation spaces \cite{tai2016towards}. It adopts an actor-critic framework that utilizes approximation functions to learn policies in continuous spaces.

The primary distinction between DDPG and D4PG lies in the latter's incorporation of the distributional Bellman operator, which is crucial for the optimization process. This operator is defined as follows:
\begin{equation}\label{distributional_bellman}
    (T_\pi Z)(s,a) = r(s,a) + \gamma\mathbb{E}\left[Z(s',\pi(s'))\ |\ s,a\right],
\end{equation}
where $Q_\pi(s,a) = \mathbb{E}\ Z_\pi(s,a)$ and $Z_\pi$ returns a distributional variable, specifically a categorical distribution. 

In the D4PG framework, the categorical distribution models the output of the critic network, which predicts a vector of probabilities over predefined reward bins, each corresponding to a range of potential reward values. This distribution allows the network to estimate a full probability distribution of expected returns, rather than a single expected value, capturing the variability in possible outcomes:
\begin{equation}
    \text{Categorical}(Z_{\pi}(s,a)) = \left[\ p_1, p_2, \dots, p_k\ \right],
\end{equation}
where $p_i$ is the probability of the return falling into the $i$-th bin and $k$ is the number of bins. The training of the critic involves minimizing the divergence between the predicted and target distributions, enhancing the policy robustness by better representing the uncertainties in dynamic environments.

Furthermore, D4PG introduces the concept of $N$-step returns to estimate the Temporal Difference (TD) error, expressed as:
\begin{equation}
\footnotesize
    \begin{aligned}
        \left(T_{\pi}^{N}Q\right)(s,a) &= r(s,a) + \\
        & \mathbb{E}\left[\sum_{n=1}^{N-1}\gamma^n r(s_n,a_n) + \gamma^N Q(s_N,\pi(s_N))\ |\ s,a\right].
    \end{aligned}
\end{equation}

In our research, we have developed an approach called \textbf{PDDRL} based on D4PG, incorporating an extension of $N$-step returns and a critic value function modeled as a categorical distribution \cite{bellemare2017distributional}. The deterministic policy is denoted by $\mu$, while noisy actions are represented by $\mu'$, where the noise process involves:
\begin{equation}
    \mu' = \mu(s_t) + \mathcal{N},
\end{equation}
and $\mathcal{N}$ follows the Ornstein-Uhlenbeck process \cite{uhlenbeck1930theory}.

Given the inconsistent experimental results reported in the original D4PG study \cite{barth2018distributed}, our research has led us to develop two variants: \textbf{PDDRL}, which utilizes classical replay memory, and \textbf{PDDRL-P}, employing prioritized replay memory \cite{schaul2015prioritized}.

Finally, a target network needs to be established to enhance learning stability. This network is a duplicate of the actor and critic networks but uses ``soft" updates. The weights $\theta'$ of the target network are gradually adjusted according to the factor $\tau$, as described by the equation:
\begin{equation}
    \theta' = \tau \theta + (1 - \tau) \theta'.
\end{equation}

\subsection{Parallel Distributional Stochastic RL}

The SAC architecture \cite{haarnoja2018soft}, proposed as a stochastic counterpart to deterministic actor-critic methods like DDPG, employs approximation functions to learn policies in continuous action spaces. SAC introduces a Bellman operator enhanced by entropy addition, aiding in exploration and policy optimization:
\begin{equation}\label{entropy_bellman}
\small
    \begin{aligned}
    (T_\pi Q)(s,a) = r(s,a) + \gamma \mathbb{E}\left[Q(s',a') - \alpha \log \pi(a' | s')\ | \ s,a\right].
    \end{aligned}
\end{equation}

To foster robust exploration, SAC emphasizes maximizing both reward and the entropy of the policy, thereby promoting action diversity and discouraging premature policy convergence. The algorithm assigns equal probability to actions yielding similar Q-values and mitigates failure risks in the Q-function approximation due to uncertain actions.

Building on this, DSAC \cite{ma2020dsac} merges the maximum entropy framework of SAC with the distributional approach of DDPG, leading to a hybrid Bellman operator:
\begin{equation}
\footnotesize
    \begin{aligned}
    (T_\pi Z)(s,a) = r(s,a) + \gamma \left[Z(s',a') - \alpha \log \pi(a' | s')\ | \ s,a\right],
    \end{aligned}
\end{equation}
which integrates both stochastic and distributional components, further described as:
\begin{equation}
\small
    \begin{aligned}
    Z_\pi(s,a) = \sum_{t=0}^{\infty} \left[r(s,a) + \gamma - \alpha \log \pi(a_{t+1} | s_{t+1})\ | \ s,a\right].
    \end{aligned}
\end{equation}

In our research, the newly proposed \textbf{PDSRL} methodology builds upon DSAC, incorporating the strategic use of N-step returns and soft updates—features previously detailed in Section~\ref{pddrl} under PDDRL—to enhance both prediction accuracy and stability.

We implemented two variants, \textbf{PDSRL} using classical replay memory, and \textbf{PDSRL-P} with prioritized replay memory, both featuring a $100000$-step sized replay memory for consistency across all Deep-RL approaches.

\subsection{Network Structure}

\begin{figure}[t]
    \centering
    \includegraphics[width=\linewidth]{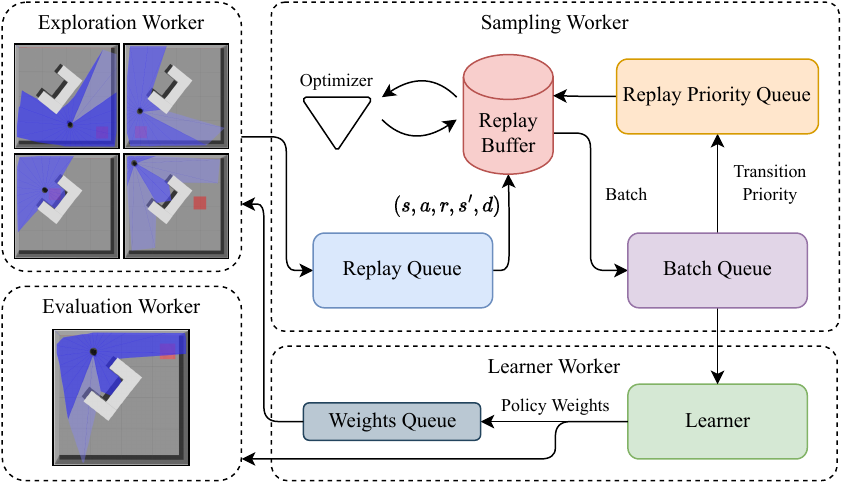}
    \caption{Parallel Deep-RL training process structure.}
    \label{fig:network_structures}
    \vspace{-6mm}
\end{figure}

All the approaches in our work employ a neural network with 26 inputs and two outputs. The inputs comprise 24 range findings of a Lidar, as well as the relative position and relative angles to the target. The sensor samples range from $0^\circ$ to $360^\circ$, and are equally spaced by $15^\circ$. The input angles serve to direct the vehicle towards the target and enhance the learning process, while the target distance encourages the network to minimize it. Meanwhile, the outputs are the linear and angular velocities that enable control of the vehicle. As a point of comparison with the Deep-RL approaches, we employed a traditional technique BBA, which uses hardcoded navigation procedures. It is important to note that the BBA uses the same amount of sensor distance information as the other approaches for a fair comparison.

The neural networks employed in all approaches have three hidden fully-connected layers, with 256 neurons each. These layers are connected through ReLU activation. The range of action is between $-1$ and $1$, and the activation function for the actor-network is the hyperbolic tangent function (\textit{Tanh}). The outputs for the linear velocity are scaled between $-0.12$ and $0.15$ meters, while the outputs for the angular velocity are scaled between $-0.1$ $m/s$ and $0.1$ $m/s$. In both approaches, the critic network predicts the Q-value of the current state, while the actor-network predicts the current state. During the training phase of the parallel approaches, four agents were employed for training, and one agent was employed for evaluation. The training process is depicted in Fig.~\ref{fig:network_structures}.

\begin{figure}[t]
    \centering
    \includegraphics[width=\linewidth]{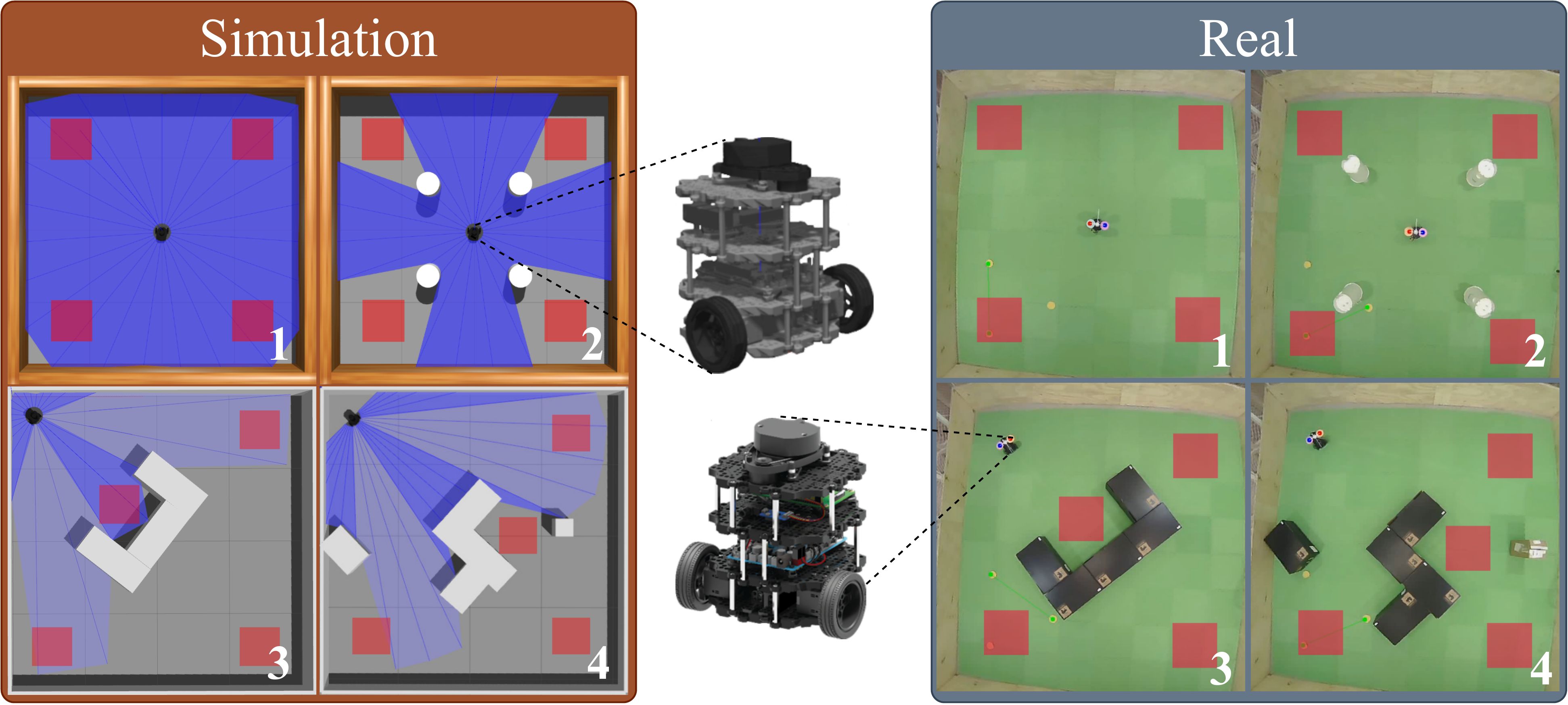}
    \caption{Simulated and real setups.}
    \label{fig:setup}
    \vspace{-6mm}
\end{figure}

\setcounter{figure}{3}
\begin{figure}[b]
\vspace{-5mm}
	\includegraphics[width=0.49\linewidth]{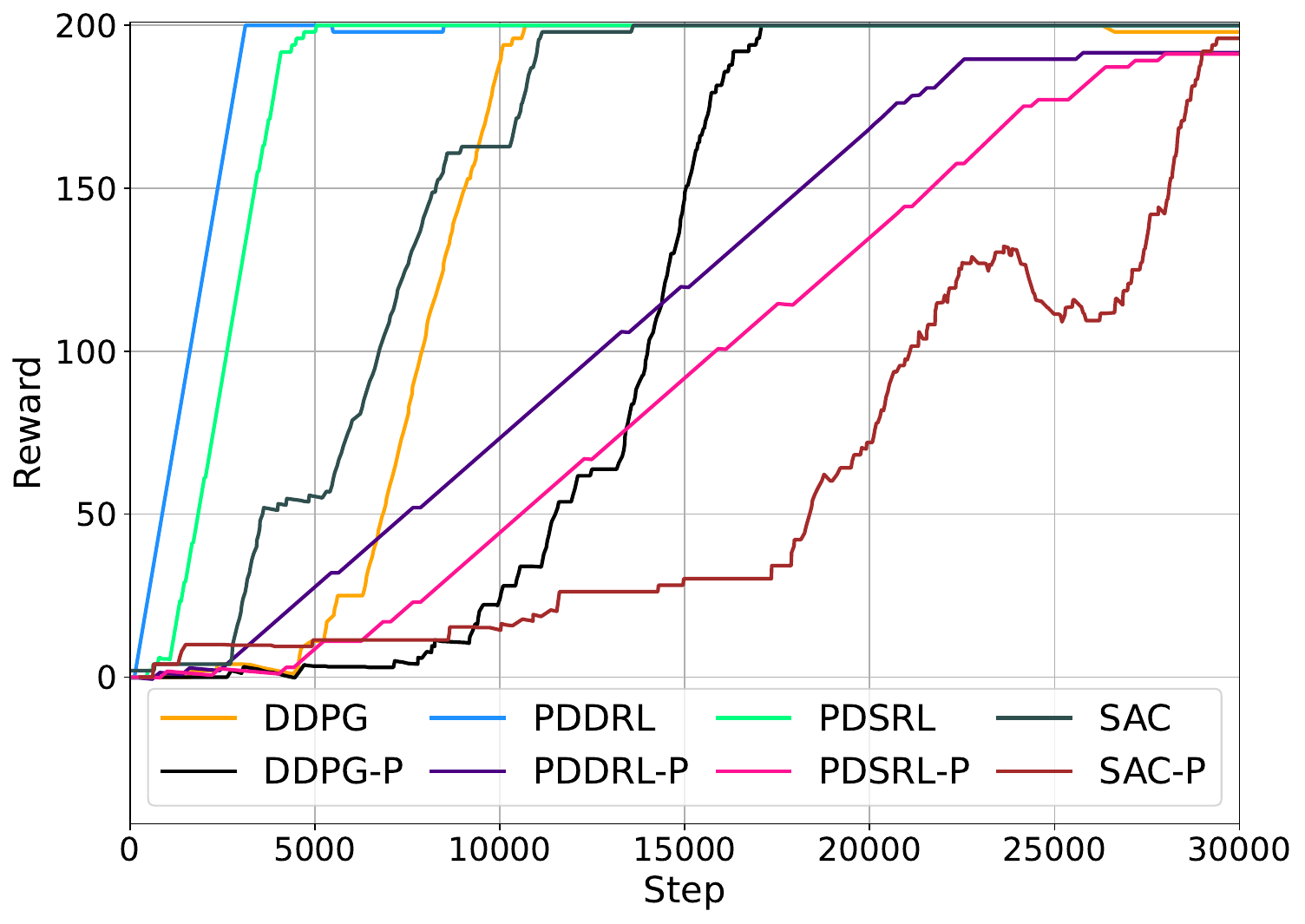}
	\includegraphics[width=0.49\linewidth]{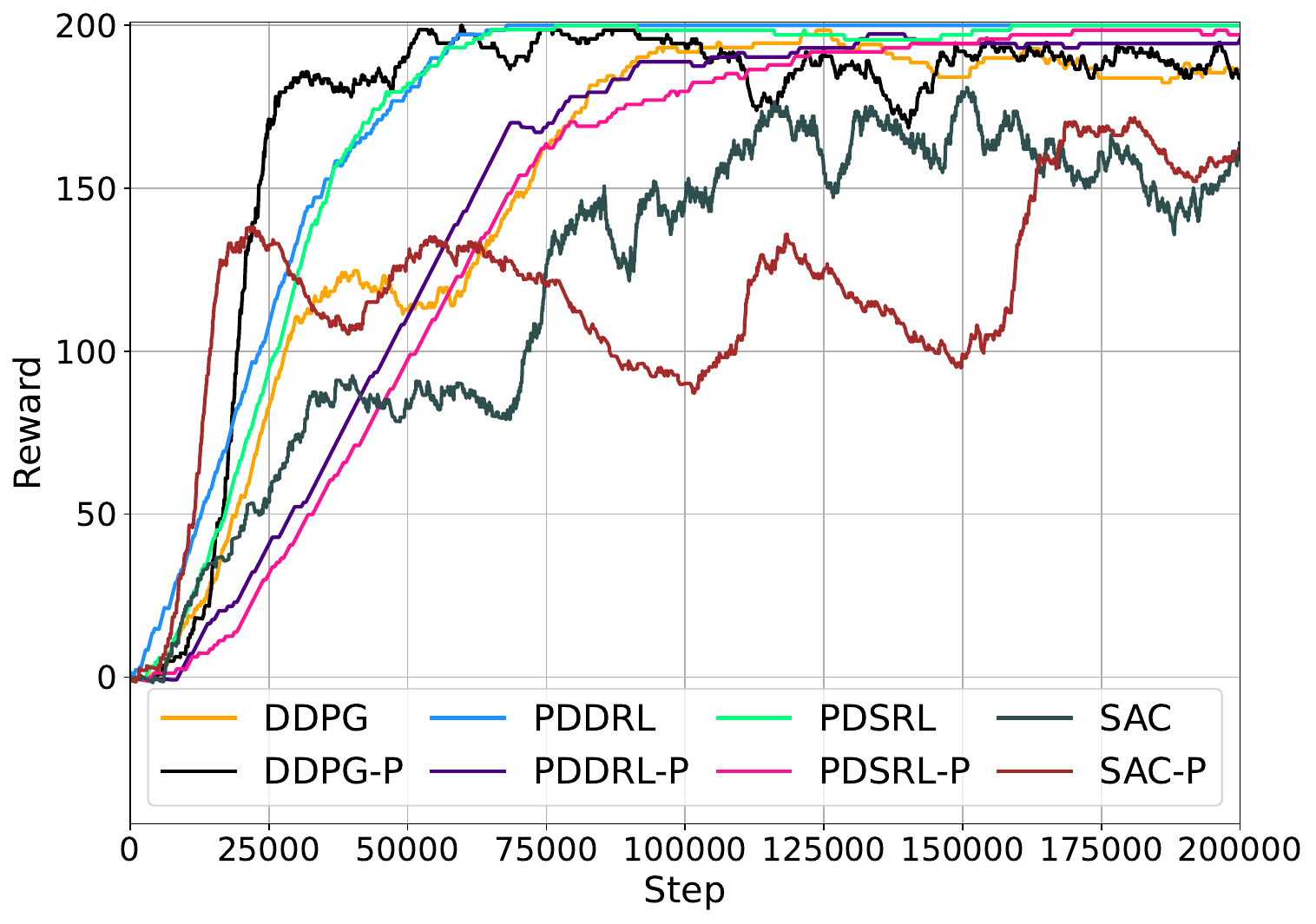}
	\includegraphics[width=0.49\linewidth]{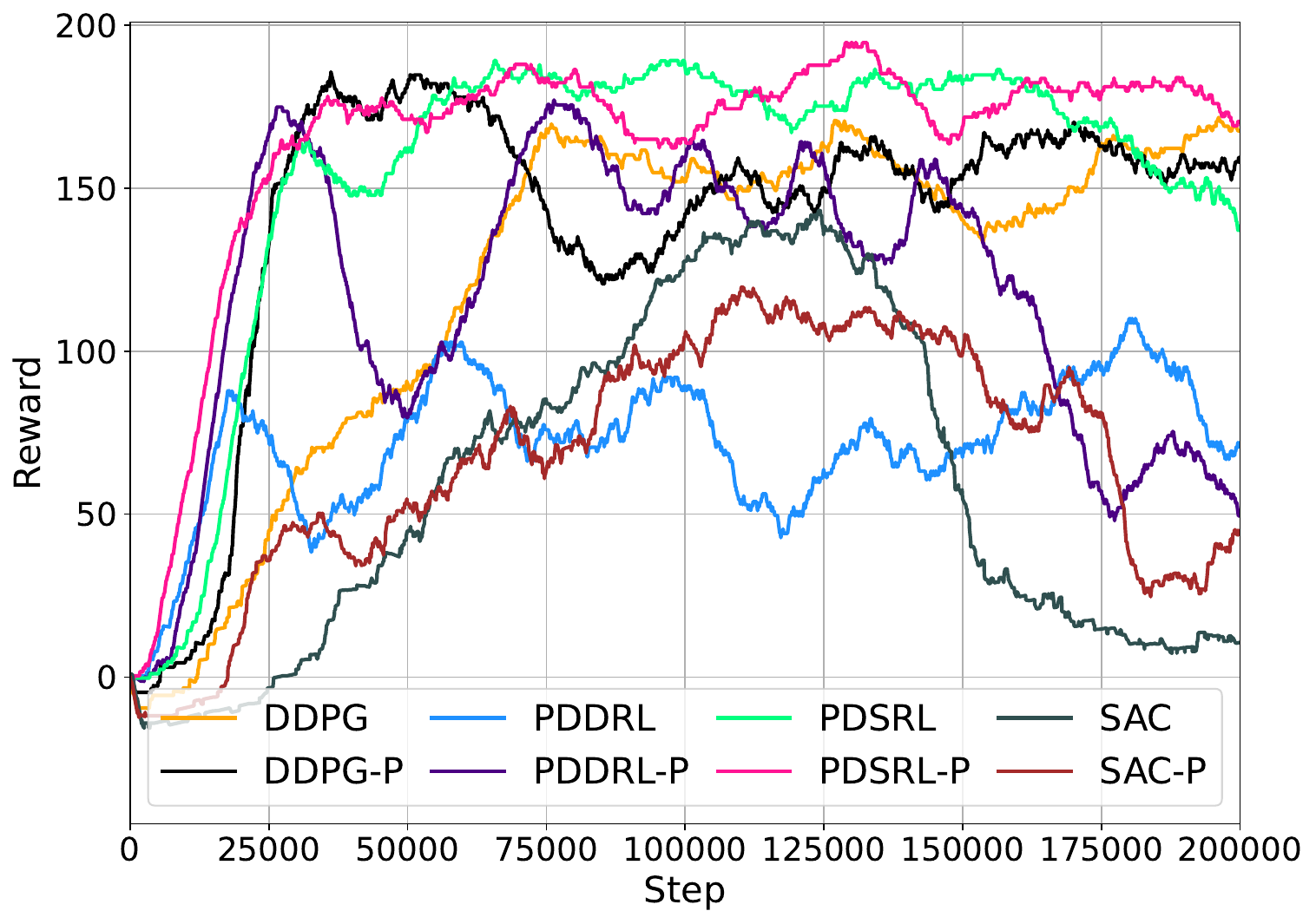}
	\includegraphics[width=0.49\linewidth]{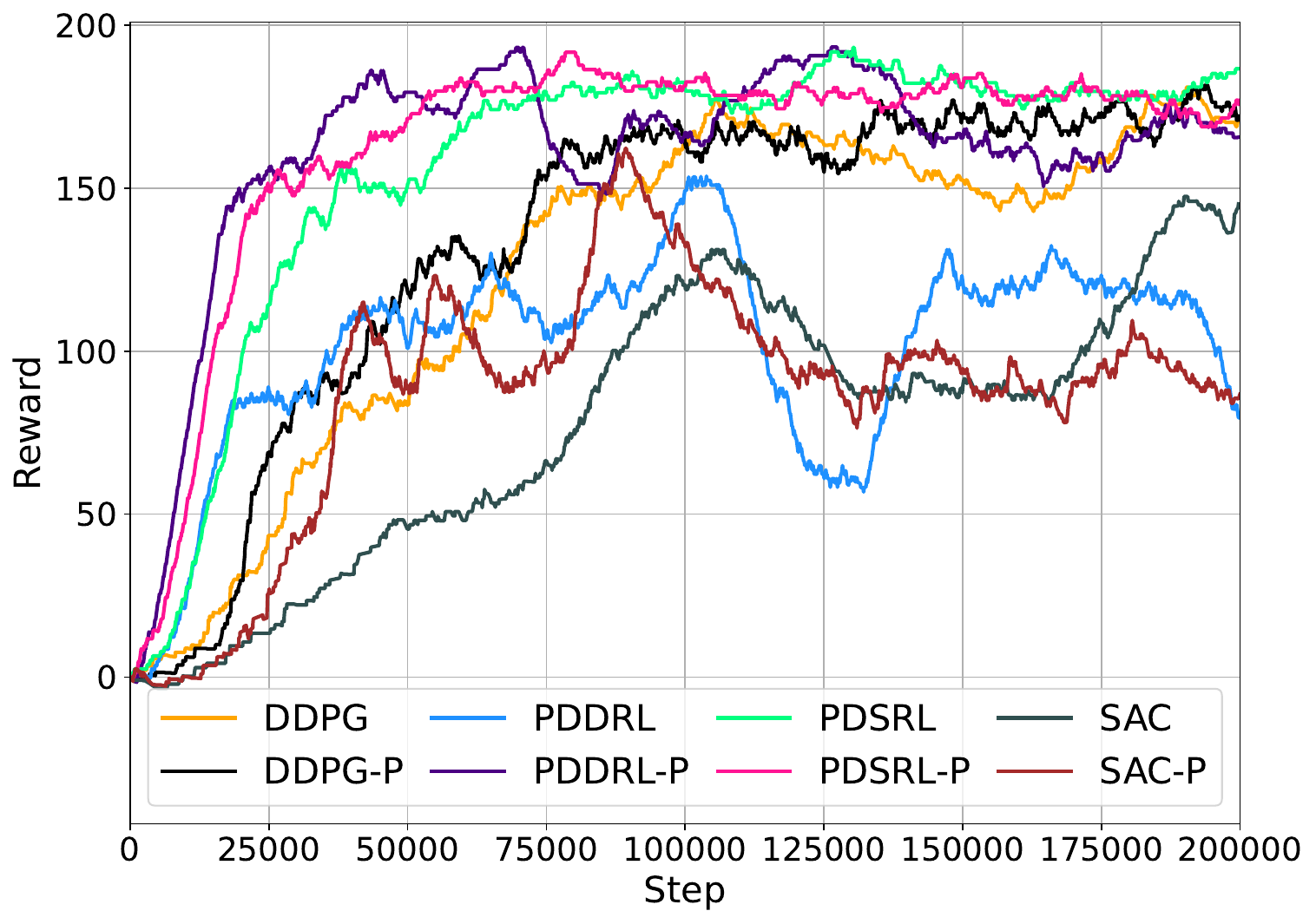}
\caption{Moving average of the agent's reward at each training step for all parallel approaches in each scenario. Scenarios organized by following the respectively order, from left to right and from top to bottom: 1, 2, 3, and 4.}
\label{fig:rewards}
\end{figure}

\setcounter{figure}{4}
\begin{figure*}[htbp]
    \centering
    \includegraphics[width=\linewidth]{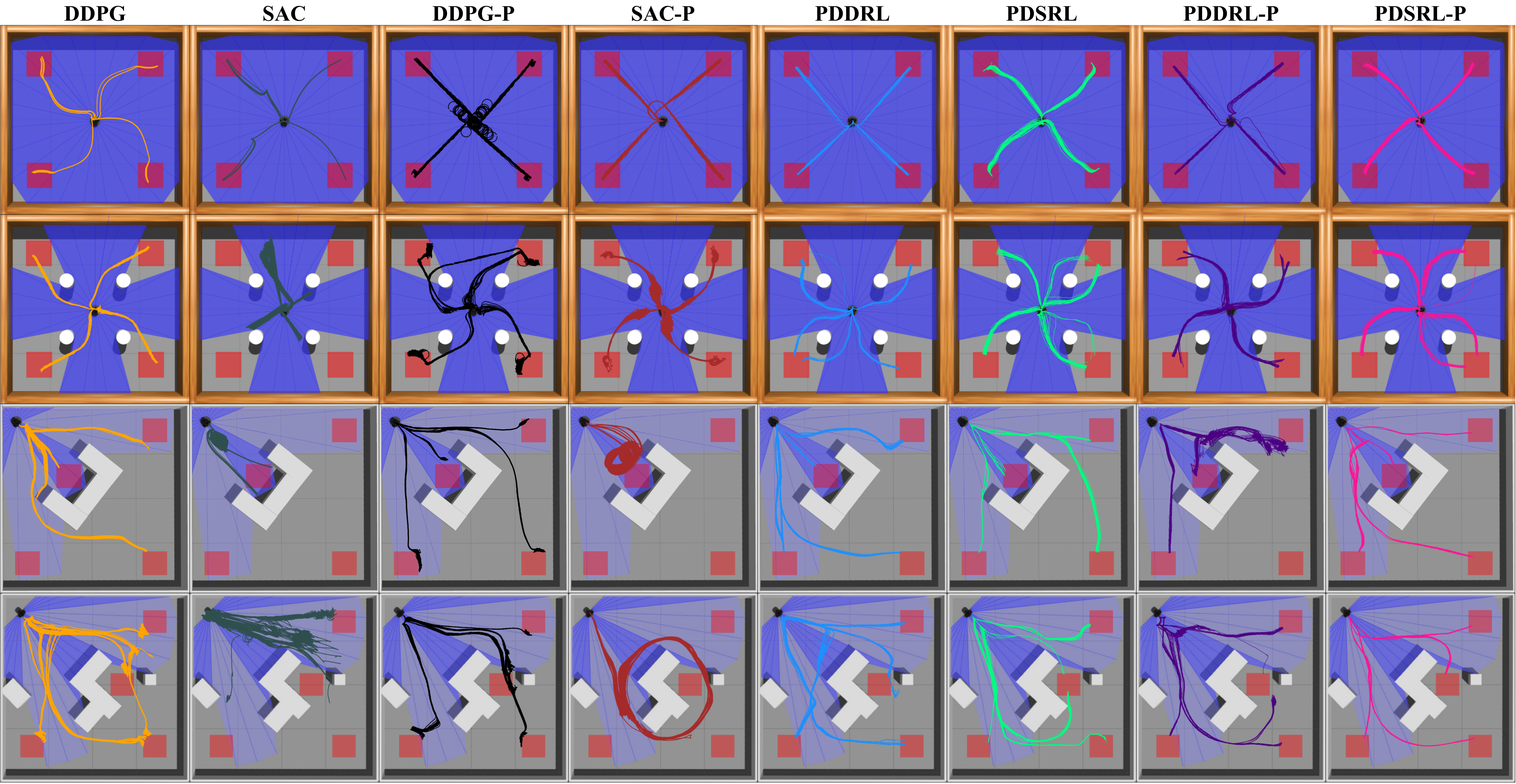}
    \caption{The behavior of each parallel approach was evaluated by conducting 100 navigation trials in each simulated scenario. The lines illustrate the paths taken by the agents, with each agent given 25 attempts to capture each target.}
    \label{fig:behavior_simulation}
\end{figure*}

\setcounter{figure}{5}
\begin{figure*}[htbp]
    \centering
    \includegraphics[width=\linewidth]{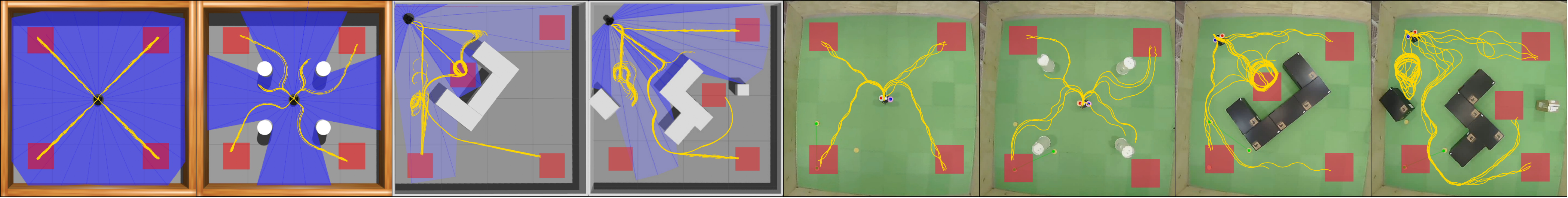}
    \caption{BBA approach evaluated in simulated and real scenarios over 100 and 12 trials, respectively. Lines represent agents' paths, with 25 (simulated) and 3 (real) attempts to capture each target.}
    \label{fig:behavior_bb}
    \vspace{-3mm}
\end{figure*}

\subsection{Reward Function}

In order to facilitate learning, it is essential to establish a reward system that encourages good actions and penalizes poor ones by the agent. This system has been designed based on empirical knowledge obtained through empirical evaluation. The reward system implemented for this work is provided below:
\begin{equation}
r (s_t, a_t) = 
\begin{cases}
r_{arr}\ \textrm{if} \ d_t < c_d\\
r_{coll}\ \textrm{if}\ min_x < c_o\\
r_{idle}\ \textrm{if}\ min_x \ge c_o\ \textrm{and}\ d_t \ge c_d
\end{cases}
\end{equation}

A simple rewarding function is defined, providing only three types of rewards: one for successfully completing the task, another for failing to complete the task, and a final one in the event that neither of the previous two occurred. A reward of $200$ is given to the agent for successfully reaching the goal within a margin of $c_d$ meters, which was set at $0.25$ meters. If the agent collides with an obstacle or reaches the limits of the scenario, a negative reward $r_{coll}$ of $-20$ is given. Collisions are detected when distance sensor readings are less than a distance of $c_o$, which is set at $0.12$ meters. If the agent's Lidar distance is greater than $c_o$ and the target distance is greater than $c_d$, a reward $r_{idle}$ of $0$ is given. This function enables a focus on the Deep-RL approaches themselves, their similarities and differences, rather than the scenario.

\section{Experiments}\label{results}

\begin{table*}[tbp]
\centering
\setlength{\tabcolsep}{6.7pt}
\caption{Precision over $100$ (for simulation) and $12$ (for real) navigation trials in four different scenarios for all approaches.}
\label{table:mean_std}
\begin{tabular}{l | c c c c c c c c c}
\mytoprule
\textbf{Scenario} & \textbf{BBA} & \textbf{DDPG} & \textbf{SAC} & \textbf{DDPG-P} & \textbf{SAC-P} & \textbf{PDDRL} & \textbf{PDSRL} & \textbf{PDDRL-P} & \textbf{PDSRL-P} \\
\mymidrule
First Sim. & $100\%$ & $100\%$ & $100\%$ & $100\%$ & $100\%$ & $100\%$ & $100\%$ & $100\%$ & $100\%$ \\
Second Sim. & $62\%$ & $100\%$ & $0.0\%$ & $85\%$ & $64\%$ & $100\%$ & $100\%$ & $100\%$ & $100\%$ \\
Third Sim. & $75\%$ & $74\%$ & $0.0\%$ & $75\%$ & $0.0\%$ & $100\%$ & $100\%$ & $70\%$ & $100\%$ \\
Fourth Sim. & $73\%$ & $81\%$ & $25\%$ & $92\%$ & $0.0\%$ & $83\%$ & $98\%$ & $76\%$ & $100\%$ \\
\mymidrule
First Real & $100\%$ & $100\%$ & $100\%$ & $100\%$ & $100\%$ & $100\%$ & $100\%$ & $100\%$ & $100\%$ \\
Second Real & $83.3\%$ & $100\%$ & $0.0\%$ & $100\%$ & $41.6\%$ & $100\%$ & $91.6\%$ & $100\%$ & $100\%$ \\
Third Real & $75\%$ & $100\%$ & $0.0\%$ & $100\%$ & $0.0\%$ & $100\%$ & $100\%$ & $0.0\%$ & $100\%$ \\
Fourth Real & $75\%$ & $66.6\%$ & $25\%$ & $100\%$ & $25\%$ & $25\%$ & $75\%$ & $75\%$ & $100\%$ \\
\mybottomrule
\end{tabular}
\vspace{-2mm}
\end{table*}

In this section, we present the experimental evaluation of our proposed approaches. The experiments are designed to assess the performance of the methods in both simulated and real-world environments. We describe the setup of the simulated and real scenarios used for training and testing, followed by a detailed analysis of the experimental results.

\subsection{Simulation and Real Scenarios}

The simulations of the robot and the scenario use the Gazebo Simulator. The connection between the robot and the agents was made using Robot Operating System (ROS). For distributing the agents, multiple instances of the Gazebo simulator are created in parallel and the data acquired are published to the replay buffer. The robot chosen here is the TurtleBot3 Burger version. The real and simulated Turtlebot3 Burger used can be seen in Fig.~\ref{fig:setup}.

Four simulation scenarios were employed in this study. The first scenario represents a navigable area for the robot to move, with the walls being the only obstacle that could cause the robot to collide. In the event of a collision with the wall or any obstacle, a negative reward is issued for the action, and the current episode is terminated. The second scenario features four fixed cylinder-shaped obstacles with a radius of 25 centimeters each. The third and fourth scenarios were built in order to create more challenging paths for the robot to reach the final goal. The third scenario includes a ``U"-shaped object that presents the agent with two possible trajectories of equal cost, with the obstacle creating a dead-end in the middle. The fourth scenario is asymmetrical and more complex, requiring the intelligent agent to develop better strategies to avoid collisions. Both real and simulated scenarios are labeled in Fig.\ref{fig:setup}.

After training the approaches through simulation, we evaluated our approaches in real scenarios. Some of the necessary data in the real scenario were obtained by image processing using OpenCV. All four real scenarios resembled the simulated ones, but their geometry is not equal. The differences of each one are more visually demonstrated in Section \ref{results}.

\subsection{Experimental Results}

We evaluated the navigation ability and spatial generalization of all approaches in both simulation and real-world scenarios. After the training phase, consisting of $30000$ steps for the first scenario and $200000$ steps for the others, we analyzed the moving average of rewards (Fig.~\ref{fig:rewards}). The results showed that stochastic distributional approaches outperformed others in all scenarios, despite a slower start. DDPG had similar reward means to the top approaches. 
Each approach was evaluated over $100$ episodes with predetermined target coordinates.

Table~\ref{table:mean_std} shows the results from the simulation experiments. Figs.~\ref{fig:behavior_simulation} and \ref{fig:behavior_bb} illustrate the behavior of Deep-RL and classical approaches. In scenario 1, all achieved $100\%$ accuracy. In more complex scenarios, distributional algorithms maintained $100\%$ accuracy, while others varied. DDPG matched the distributional algorithms.

\begin{table}[bp]
\vspace{-5mm}
\centering
\setlength{\tabcolsep}{11.5pt}
\caption{Precision and distance metrics over 12 navigation trials in the extra scenario for top approaches.}
\label{table:extra_scenario}
\begin{tabular}{l | c c}
\mytoprule
\textbf{Algorithm} & \textbf{Precision (\%)} & \textbf{Crashes (\%)} \\
\mymidrule
\textbf{DDPG} & $58.3\%$ & $41.6\%$ \\
\textbf{DDPG-P} & $100\%$ & $0.0\%$ \\
\textbf{PDDRL} & $100\%$ & $0.0\%$ \\
\textbf{PDSRL} & $75\%$ & $8.3\%$ \\
\textbf{PDSRL-P} & $100\%$ & $0.0\%$ \\
\mybottomrule
\end{tabular}
\end{table}

In Scenario 3, distributional approaches outperformed others, except for PDDRL-P, which had $70\%$ accuracy. This issue is known in the literature, where prioritized memory may limit the generalization capacity of the deterministic distributional agent based on the D4PG algorithm. In Scenario 4, only the PDSRL-P approach captured all points, followed by PDSRL with $98\%$ accuracy. Notably, DDPG-P achieved $92\%$ accuracy, surpassing deterministic distributional algorithms.

Fig.~\ref{fig:behavior_simulation} shows PDSRL and PDSRL-P had smoother trajectories. DDPG showed instability with repeated collisions in Scenario 3. Prioritization reduced collisions in Scenario 4, but instability remained.
Real environment experiments used target points similar to simulations. Fig.~\ref{fig:behavior_real} shows trajectories in all real scenarios. PDSRL-P and DDPG-P achieved $100\%$ accuracy in real environments. Performance differences between PDDRL and PDDRL-P were significant in scenarios 3 and 4. Real scenarios introduced delays, affecting performance.

\setcounter{figure}{6}
\begin{figure*}[htbp]
    \centering
    \includegraphics[width=0.85\linewidth]{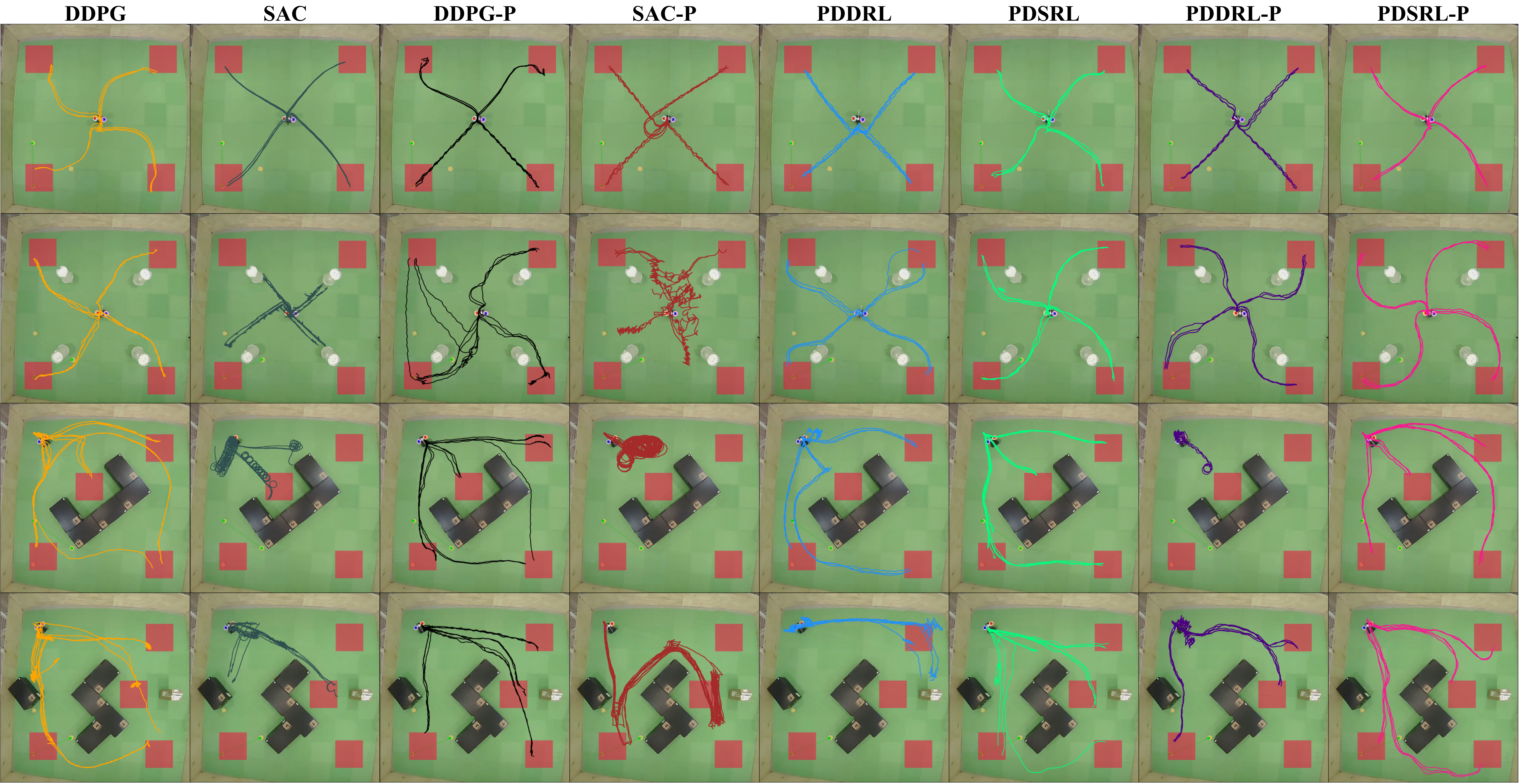}\\
    \includegraphics[width=0.131\textwidth]{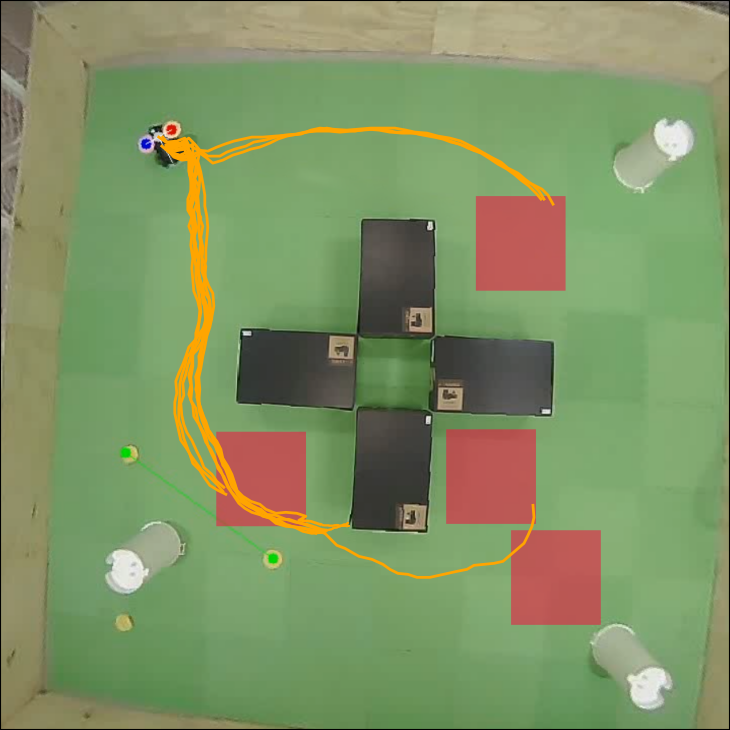}
    \includegraphics[width=0.131\textwidth]{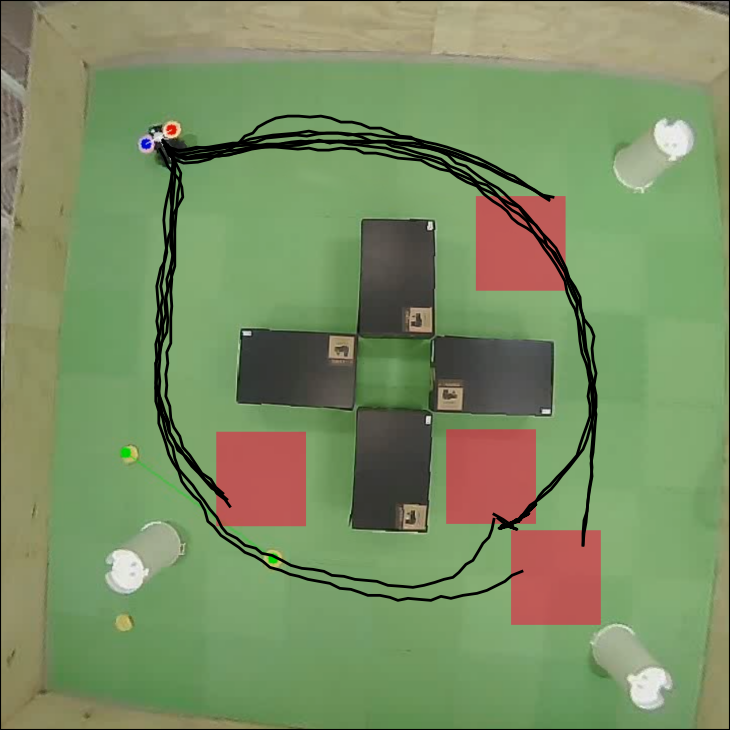}
    \includegraphics[width=0.131\textwidth]{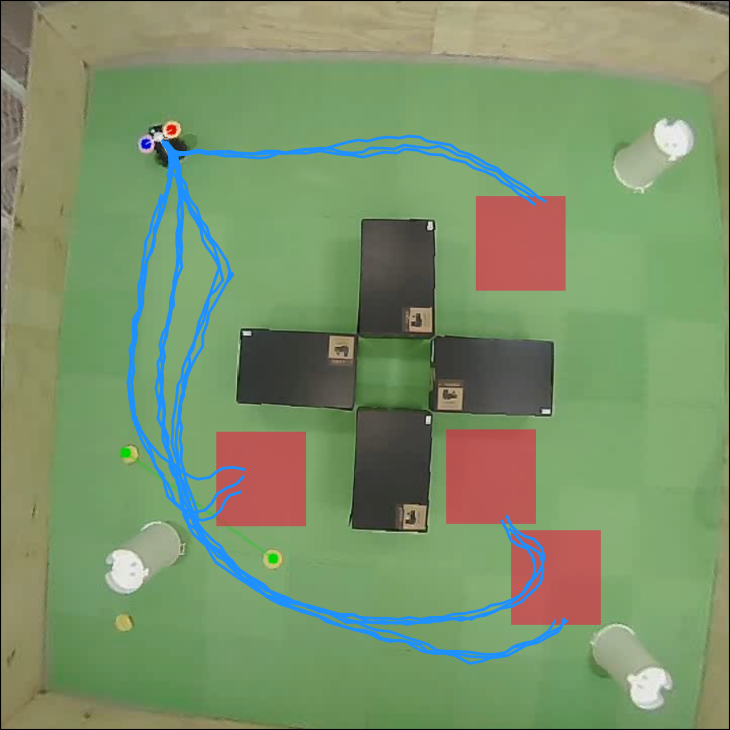}
    \includegraphics[width=0.131\textwidth]{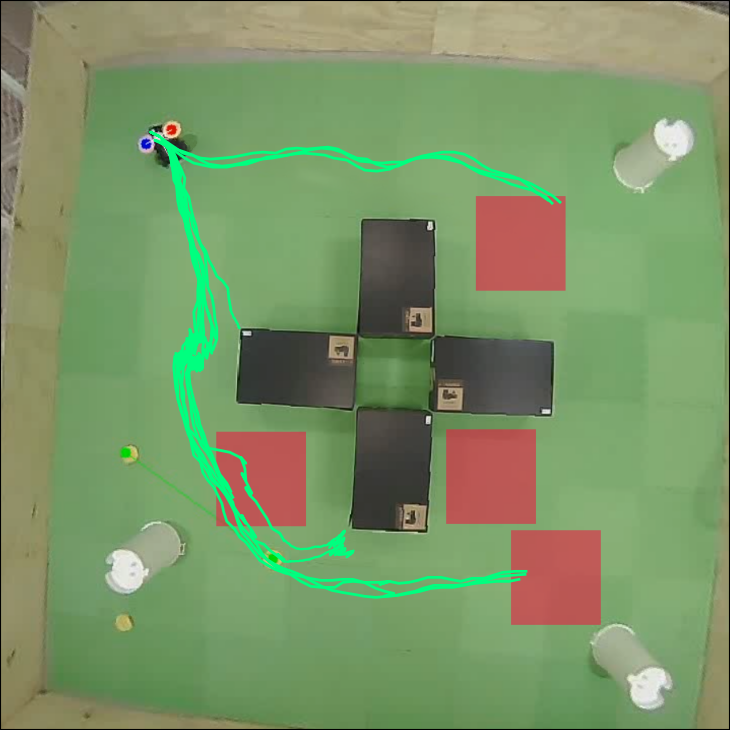}
    \includegraphics[width=0.131\textwidth]{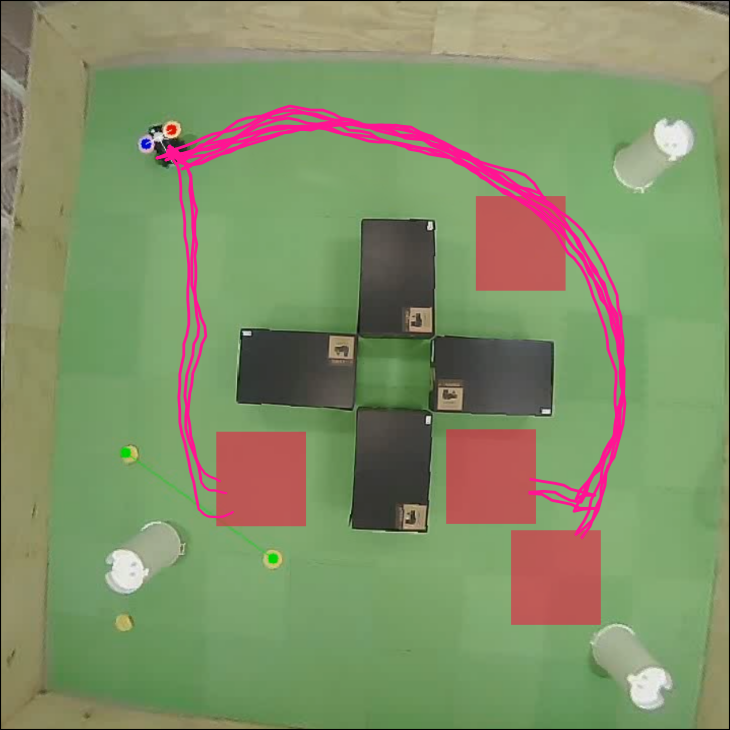}
    \caption{(Top) The behavior of each parallel approach was evaluated by conducting 12 navigation trials in each real scenario. The lines illustrate the paths taken by the agents, with each agent given 3 attempts to capture each target. (Bottom) The behavior of the best classified parallel approaches evaluated in the extra real scenario. Organized by the following order: DDPG, DDPG-P, PDDRL, PDSRL, and PDSRL-P.}
    \label{fig:behavior_real}
    \vspace{-3mm}
\end{figure*}

We proposed a novel scenario where pre-trained models from scenario 3 were evaluated in an unseen environment. Fig.~\ref{fig:behavior_real} and Table~\ref{table:extra_scenario} show the results. DDPG had high crash rates, while PDSRL minimized crashes but didn't always achieve rewards. DDPG-P, PDDRL, and PDSRL-P achieved $100\%$ accuracy, demonstrating strong spatial generalization.

Our extensive validation shows the proposed approaches perform well in all scenarios, including the novel one. Deep-RL approaches outperformed traditional algorithms in real-world challenges. Prioritized versions achieved the best success rates, particularly PDSRL-P. Evaluation in a novel scenario confirmed the approaches' learning capabilities and spatial generalization.

\section{Conclusion}\label{conclusions}

This work introduces novel Deep Reinforcement Learning (Deep-RL) techniques for terrestrial mobile robots to perform mapless navigation, demonstrating their learning capabilities in simulation and potential for real-world implementation. The results highlight that parallel distributional Deep-RL approaches can effectively tackle complex real-world robotics problems that non-learning-based and non-distributional approaches without prioritized memory replay cannot address.

%
\bibliographystyle{IEEEtran}
\bibliography{ref}
%

\end{document}